\documentclass{INTERSPEECH2023}
\usepackage{amsmath,graphicx,amsfonts,amssymb,multirow,footmisc,xurl}

\interspeechcameraready

\title{LAMASSU: A Streaming Language-Agnostic Multilingual Speech Recognition and Translation Model Using Neural Transducers}
\name{
\begin{tabular}{c} Peidong Wang, Eric Sun, Jian Xue, Yu Wu, Long Zhou, \\ Yashesh Gaur, Shujie Liu, Jinyu Li
\end{tabular}}
\address{Microsoft}
\email{\{peidongwang, ersun, jiaxue, wu.yu, long.zhou, yashesh.gaur, shujliu, jinyli\}@microsoft.com}

\begin{document}

\maketitle
 
\begin{abstract}
Automatic speech recognition (ASR) and speech translation (ST) can both use neural transducers as the model structure. It is thus possible to use a single transducer model to perform both tasks. In real-world applications, such joint ASR and ST models may need to be streaming and do not require source language identification (i.e. language-agnostic). In this paper, we propose LAMASSU, a streaming language-agnostic multilingual speech recognition and translation model using neural transducers. Based on the transducer model structure, we propose four methods, a unified joint and prediction network for multilingual output, a clustered multilingual encoder, target language identification for encoder, and connectionist temporal classification regularization. Experimental results show that LAMASSU not only drastically reduces the model size but also reaches the performances of monolingual ASR and bilingual ST models.
\end{abstract}
\noindent\textbf{Index Terms}: streaming, language-agnostic, multilingual, automatic speech recognition, speech translation

\section{Introduction}
\label{sec:intro}
Automatic speech recognition (ASR) may be viewed as a special case of speech translation (ST), where source and target languages are the same. When ASR and ST are both formulated as end-to-end (E2E) tasks, they can use the same model structure. Weiss \emph{et al.} first showed that an attention-based E2E (AED) ASR model \cite{watanabe2017hybrid,chiu2018state,he2019streaming,Li2020comparison,wang2019token,wang2021mute} can be used for ST \cite{weiss2017sequence}. Recently, neural transducer \cite{Graves-RNNSeqTransduction,prabhavalkar-comparison,E2EOverview}, another popular E2E ASR model structure, is applied to streaming ST tasks on large-scale weakly supervised datasets \cite{xue2022large}.

The convergence of model structures for ASR and ST inspires works that use a single model to perform both ASR and ST 
\cite{wang2020covost,salesky2021multilingual,sperber2020consistent,karakanta2021between,xu2022joint}
. Liu \emph{et al.} proposed an interactive decoding strategy between ASR and ST \cite{liu2020synchronous}. Le \emph{et al.} proposed dual-decoder Transformer for joint ASR and multilingual ST \cite{le2020dual}, which is evaluated on a one-to-many ST dataset.  Radford \emph{et al.} proposed Whisper \cite{radford2022robust}, which performs multilingual ASR and many-to-one ST in a single offline AED model.
Some recent works on speech pre-training are evaluated on both ASR and ST tasks \cite{tang2021fst,tang2022unified,bapna2022mslam,chen2022maestro}. 

To be useful for real-world conversations, we think a joint ASR and ST model should be streaming, multilingual, and agnostic to source languages. Therefore, we propose a streaming \textbf{l}anguage-\textbf{a}gnostic \textbf{m}ultilingu\textbf{a}l \textbf{s}peech recognition and tran\textbf{s}lation model using neural transd\textbf{u}cers (LAMASSU) in this paper. Transformer transducer (TT) \cite{xiechen2021tt} is leveraged as the backbone network to perform streaming ASR and ST. The main contributions of this paper are:

\begin{enumerate}
    \item This is the first work using a neural transducer as a single model for multilingual ASR and many-to-many ST.
    \item We adopt the multilingual decoder in AED-based ST models \cite{inaguma2019multilingual,li2021multilingual} for neural transducers.
    \item We propose a novel clustered multilingual speech encoder for LAMASSU \cite{wang2019speech,wang2020speaker},
    enabling the model to perform multilingual ASR and ST in code-switch scenarios \cite{weller2022end,huber2022code}.
    \item We propose to feed target language identification (LID) to encoder to improve model performances \cite{elnokrashy2022language}.
    \item We apply an auxiliary connectionist temporal classification (CTC) loss \cite{zhang2022revisiting,watanabe2018espnet,yan2022ctc} to regularize the training procedure for transducer-based ST models.
    
\end{enumerate}
Experimental results show that with only 1/5 of model parameters, LAMASSU can reach the performance of baseline monolingual ASR and bilingual ST models.

\begin{figure}[t]
    \centering
    \centerline{\includegraphics[width=.6\linewidth]{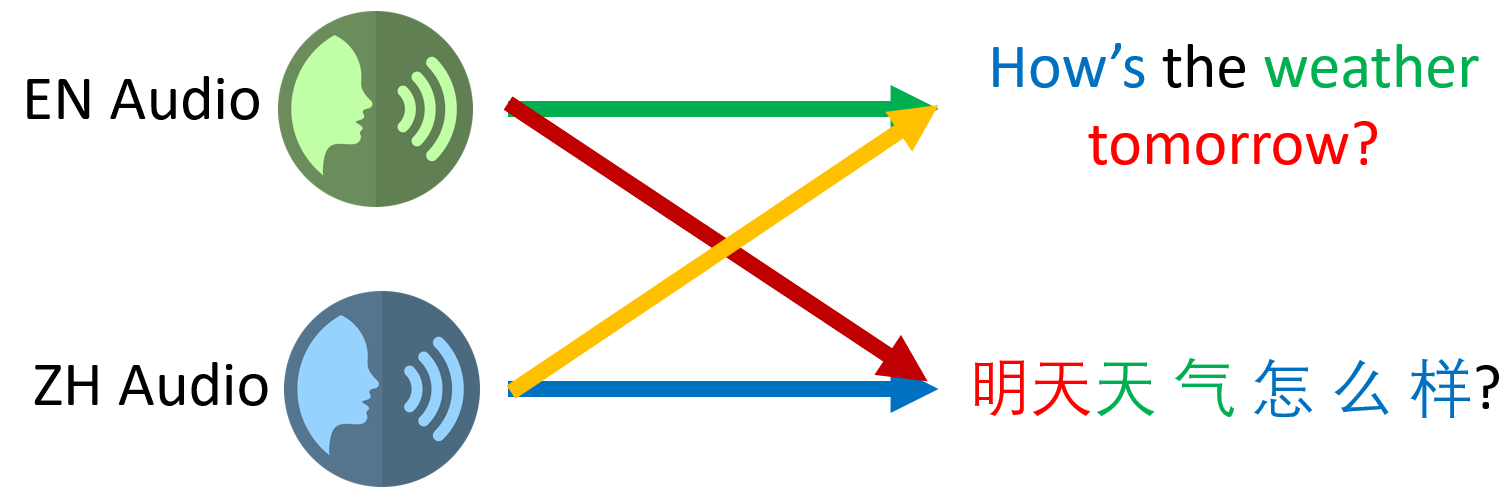}}
	\caption{Illustration of a single model for multilingual ASR and many-to-many ST.}
	\label{fig:multilingual_asr_st}
\end{figure}

The remainder of this paper is organized as follows. We describe LAMASSU in Section \ref{sec:sys}. In Section \ref{sec:exp} and \ref{sec:eval}, we show experimental setup and evaluation results, respectively. We conclude in Section \ref{sec:conclusion}.

\section{LAMASSU}
\label{sec:sys}
LAMASSU is based on the neural transducer model structure, which contains a prediction network, a joint network, and an encoder. To enable multilingual output, we first design multilingual prediction and joint networks. Then we propose a clustered multilingual encoder for LAMASSU. After the illustration of the proposed model structure in the first two subsections, we introduce the use of target LID for encoders, and CTC regularization for transducer training.


\subsection{Multilingual Prediction and Joint Networks}
\label{ssec:multilingual_st}
There are mainly two ways to generate texts of multiple target languages: specified prediction and joint networks for different languages, or a unified prediction and joint network selecting output languages using target LIDs.
Specified prediction and joint networks are more flexible during deployment and when extended to new target languages, whereas a unified prediction and joint network may leverage the shared information of different language pairs. 

\subsubsection{LAMASSU-SPE}
\label{sssec:specified}

\begin{figure}[htb]
    \centering
    \centerline{\includegraphics[width=.8\linewidth]{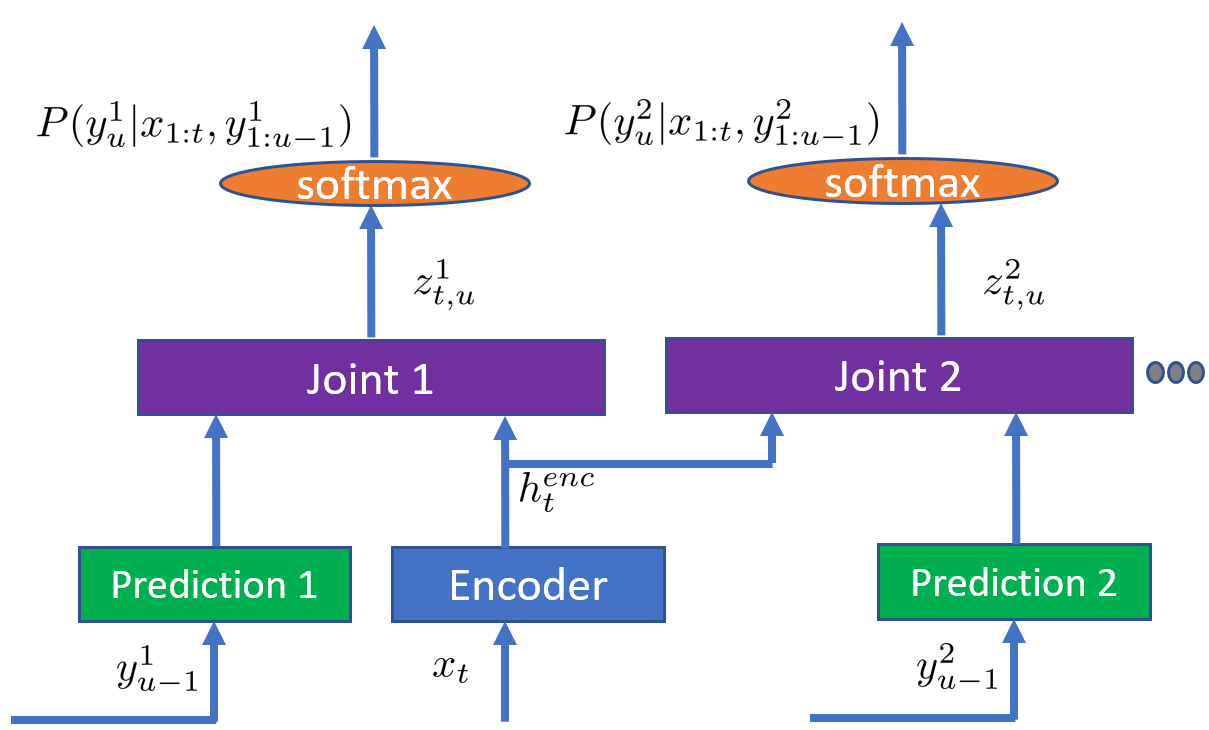}}
	\caption{Illustration of LAMASSU-SPE. It uses specified prediction and joint networks for different target languages. For notation simplicity, we use the same time scale $t$ for $\mathbf{x}_t$ and $\mathbf{h}^\mathrm{enc}_t$. In practice, there is typically a subsampling layer in the encoder.}
	\label{fig:transducer_separate_decoders}
\end{figure}

Fig. \ref{fig:transducer_separate_decoders} illustrates LAMASSU using specified prediction and joint networks for different target languages (LAMASSU-SPE). 
An encoder converts input feature $\mathbf{x}_t \in \mathbb{R}^{d_x}$ to a hidden representation $\mathbf{h}^\mathrm{enc}_t \in \mathbb{R}^{d_h}$. The input to prediction network $k$ is the decoding result $\mathbf{y}^k_{u-1} \in \mathbb{R}^1$ at output step $u-1$, where $k$ is an index of the target language. Joint network $k$ combines $\mathbf{h}^\mathrm{enc}_t$ and the output of prediction network $k$ to get the output $\mathbf{z}^k_{t,u} \in \mathbb{R}^{d_z^k}$ at $t$ and $u$. After a $\mathrm{softmax}$ operation, the network generates the probability $P(\mathbf{y}^k_u \in \mathbf{Y}^k \cup \varnothing | \mathbf{x}_{1:t}, \mathbf{y}^k_{1:u-1})$ for output token at step $u$, where $\mathbf{Y}^k$ denotes the vocabulary list for target language $k$ and blank token $\varnothing$ is used to align the input and output steps.


\subsubsection{LAMASSU-UNI}
\label{sssec:unified}
\begin{figure}[h]
    \centering
    \centerline{\includegraphics[width=.6\linewidth]{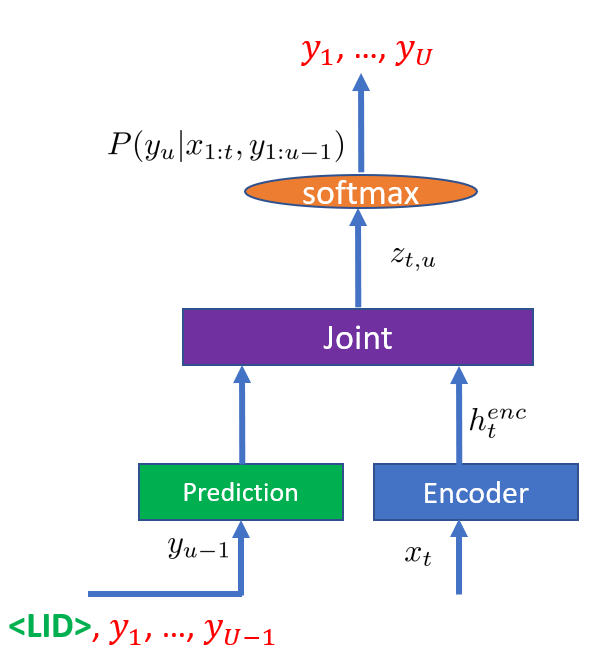}}
	\caption{Illustration of LAMASSU-UNI. It uses a unified prediction and joint network and specifies target language with the first input token to the prediction network.}
	\label{fig:transducer_single_decoder}
\end{figure}

Fig. \ref{fig:transducer_single_decoder} shows LAMASSU using a unified prediction and joint network for multiple target languages (LAMASSU-UNI). During training, it uses target LID to replace the start of sentence token ($<SOS>$) in the input to the prediction network. At test time, target LID is fed to the prediction network as the initial token.
To prepare the vocabulary list for LAMASSU-UNI, we merge the vocabularies of all supported target languages.




\subsection{Clustered Multilingual Encoder}
\label{ssec:multilingual_enc}
A straightforward way to perform joint multilingual ASR and many-to-many ST is to use a shared encoder for all source languages. However, different source languages may interfere with each other. We propose a clustered multilingual encoder, aiming to use separate Transformer modules for different clusters of languages. It is achieved by a specifically designed training scheme guided by source LIDs. Note that at inference time, no source LID information is used (i.e. language-agnostic). The shared encoder may be viewed as a special case of the clustered multilingual encoder, where the number of clusters is 1.



\subsubsection{Model Structure}

\begin{figure}[htb]
    \centering
    \centerline{\includegraphics[width=.9\linewidth]{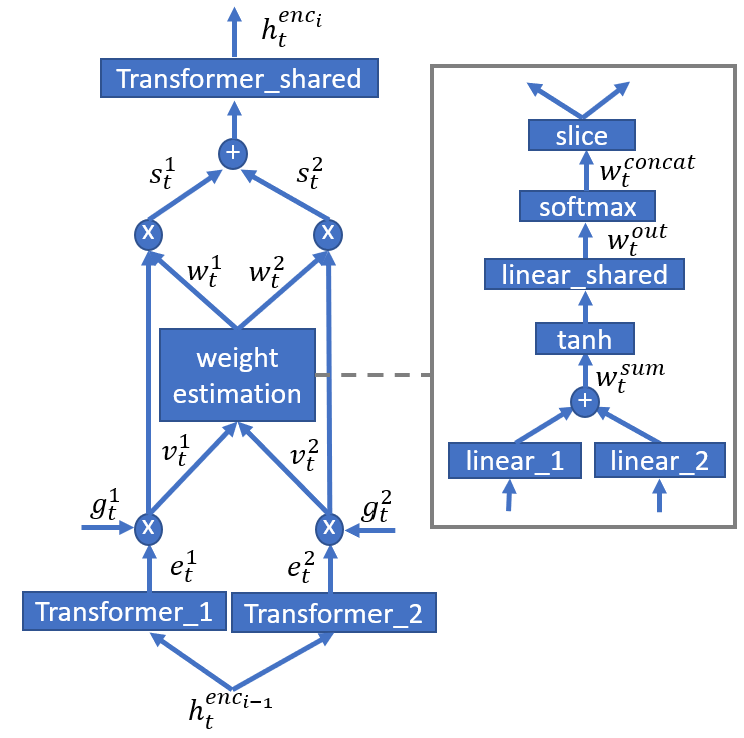}}
	\caption{Illustration of the $i$th block in a clustered multilingual encoder.}
	\label{fig:multilingual_encoder}
\end{figure}

\label{sssec:multilingual_encoder_structure}
Fig. \ref{fig:multilingual_encoder} shows the $i$th block of a clustered multilingual encoder. 
The output of the ($i-1$)th block $\mathbf{h}^{enc_{i-1}}_t$ is first fed to separate Transformer modules for different language clusters. In this paper, we only use two source languages and two separate Transformer modules due to computational limitations. In practice, we can increase the number of source languages, cluster some of them, treat the languages in each cluster as one language, and adjust the number of separate Transformer modules based on the budget.

The output of separate Transformer modules $\mathbf{e}^j_t$'s are gated by external input values $\mathbf{g}^j_t$'s to generate $\mathbf{v}^j_t$'s, where $j \in [1, J]$ corresponds to language clusters. The details of gating values $\mathbf{g}^j_t$'s are shown in the next subsection \ref{sssec:learning_scheme}.
\begin{equation}
    \mathbf{v}^j_t = \mathbf{e}^j_t * \mathbf{g}^j_t
\end{equation}
The gated values $\mathbf{v}^j_t$'s are fed to a weight estimation module shown in the gray box of Fig. \ref{fig:multilingual_encoder}. 
In the weight estimation module, each $\mathbf{v}^j_t$ is first projected using a linear layer and then summed to get $\mathbf{w}^{sum}_t$.
\begin{equation}
    \mathbf{w}^{sum}_t = \sum_{j=1}^J \mathrm{linear}_j(\mathbf{v}^j_t)
\end{equation}
We then apply $\mathrm{tanh}$ and a shared linear layer to $\mathbf{w}^{sum}_t$ to get $\mathbf{w}^{out}_t$. 
\begin{equation}
    \mathbf{w}^{out}_t = \mathrm{linear}_{\mathrm{shared}}(\mathrm{tanh}(\mathbf{w}^{sum}_t))
\end{equation}
Note that the output dimension of the shared linear layer is $J$, the same as the total number of language clusters. 
We apply a $\mathrm{softmax}$ operation to $\mathbf{w}^{out}_t$ so that the values are normalized and they sum to one.
\begin{equation}
    \mathbf{w}^{concat}_t = \mathrm{softmax}(\mathbf{w}^{out}_t)
\end{equation}
We slice the $J$-dimension weight estimation $\mathbf{w}^{concat}_t$ to obtain estimated weights $\mathbf{w}^j_t$ for each language cluster.
\begin{equation}
    \mathbf{w}^1_t, \mathbf{w}^2_t, ..., \mathbf{w}^J_t = \mathrm{slice}(\mathbf{w}^{concat}_t)
\end{equation}
After weight estimation, we multiply each weight $\mathbf{w}^j_t$ with its corresponding $\mathbf{v}^j_t$ to get $\mathbf{s}^j_t$.
\begin{equation}
    \mathbf{s}^j_t = \mathbf{w}^j_t * \mathbf{v}^j_t
\end{equation}
The summation of $\mathbf{s}^j_t$'s is fed to a shared Transformer module to obtain the final output $\mathbf{h}^{enc_i}_t$.
\begin{equation}
    \mathbf{h}^{enc_i}_t = \mathrm{Transformer}_{\mathrm{shared}}(\sum_{j=1}^J \mathbf{s}^j_t)
\end{equation}

\subsubsection{Source LID Scheduling for Training}
\label{sssec:learning_scheme}
During training, we leverage source LIDs as gating value $\mathbf{g}^j_t$'s. The training procedure is divided into two phases. In phase 1, only the $\mathbf{g}^j_t$ corresponding to the source language cluster is set to $1$, and all other $\mathbf{g}^j_t$'s are set to $0$'s. After a manually set number of training steps, we enter phase 2 and set all $\mathbf{g}^j_t$'s to $1$'s, regardless of the source language.

To better regularize the training procedure for separate Transformer modules, we add an LID loss $\mathbb{L}_{LID}$. If we denote the $\mathbf{w}^{out}_t$ in block $i$ as $\mathbf{w}^{out_i}_t$, $\mathbb{L}_{LID}$ can be calculated as a cross-entropy loss between the summation of $\mathbf{w}^{out_i}_t$'s in different blocks and the source LID:
\begin{equation}
    \mathbb{L}_{LID} = \mathrm{cross\_entropy}(\sum_{i=1}^I \mathbf{w}^{out_i}_t, j)
\end{equation}
where $I$ is the total number of blocks in the encoder.

At test time, we set all $\mathbf{g}^j_t$'s to $1$'s so that the model does not need source LID and is thus agnostic to source languages.



\subsection{Target LID for Encoders}
\label{ssec:lid_enc}
Feeding target LID to encoders may help the model to perform similarly as a many-to-one model, alleviating the interference among different language pairs. This is similar to domain identification information commonly used in robust ASR \cite{wang2018utterance,wang2019bridging,wang2019enhanced}.
To apply target LID for encoders, we append a one-hot vector corresponding to target language $k$ to each frame of input feature $\mathbf{x}$ during training and test.

\subsection{CTC Regularization}
\label{ssec:ctc_reg}
CTC regularization may be necessary when the size of training set is small.
To regularize the training procedure for transducers, we add an auxiliary CTC loss. It uses the same labels as those for the transducer loss. To generate the CTC output, we reuse the joint networks and thus do not introduce additional parameters to the model. More specifically, if we denote the joint network in a transducer as
\begin{equation}
    \mathbf{z}_{t,u} = \mathrm{linear_{out}}(\mathrm{tanh}(\mathrm{linear_{enc}}( \mathbf{h}_t^{\mathrm{enc}}) + \mathrm{linear_{pred}}(\mathbf{h}_u^{\mathrm{pred}})))
    \label{eq:joint_network}
\end{equation}
where $\mathbf{h}_u^{\mathrm{pred}}$ is the prediction network output at step $u$. The CTC output $\mathbf{c}_t$ is generated by removing $\mathrm{linear_{pred}}(\mathbf{h}_u^{\mathrm{pred}})$.
\begin{equation}
    \mathbf{c}_t = \mathrm{linear_{out}}(\mathrm{tanh}(\mathrm{linear_{enc}}( \mathbf{h}_t^{\mathrm{enc}})))
\end{equation}
Then we calculate the CTC loss $\mathbb{L}_{CTC}$ between $\mathbf{c}_t$ and the translation labels.
Note that at each training step for LAMASSU-SPE, there is only one target language $k$, and thus we only generate one $\mathbf{c}_t^k$ accordingly.

The overall loss function used for LAMASSU can be written as Equation (\ref{eq:ctc_loss}) below:
\begin{equation}
    \mathbb{L} = \mathbb{L}_{transducer} + \alpha \mathbb{L}_{LID} + \beta \mathbb{L}_{CTC}
    \label{eq:ctc_loss}
\end{equation}
where $\alpha$ and $\beta$ differ for different LAMASSU variants. If we use the clustered multilingual encoder, $\alpha$ is set to $0.75$ empirically, otherwise it is $0$. $\beta$ is set to $0.4$ when we use CTC regularization.

\section{Experimental Setup}
\label{sec:exp}
For the training data in this study, we used 10 thousand (K) hours of in-house English (EN) audio and 10K hours of in-house German (DE) audio data. 
For each of the source languages EN and DE, we obtained the transcription or translation labels for three output languages: EN, DE, and Chinese (ZH). The translation labels were generated using a text machine translation model. 
Note that our data is an order of magnitude larger than most open-source datasets \cite{wang2020covost,must-c,ye2022gigast} and we could not find any existing dataset that satisfies our experimental requirement.
To evaluate multilingual ASR and ST performances, we used publicly available test sets. Note that we use in-house training data for product purposes, which is closer to real application scenarios but is mismatched with public test sets. For EN and DE ASR, we used Librispeech and the first 2000 utterances in common voice cv-corpus-11.0-2022-09-21 DE test set. For EN-DE, EN-ZH, and DE-EN ST sets, we adopted MSLT v1.0, MSLT v1.1, and CoVoST2 DE-EN test sets. We did not evaluate bilingual evaluation understudy (BLEU) scores on DE-ZH due to the lack of high quality professionally translated test data, but recorded a one-minute demo showing LAMASSU's streaming DE-ZH translation ability  \footnote{\url{https://youtu.be/ZF_48yuBo9M?t=1860} \label{cvpr_video}}.


For the model configurations of monolingual ASR and bilingual ST, we used TT models with a chunk size of 160ms. We set the number of encoder layers to 24. For the clustered multilingual encoder, we used 6 blocks, each of which is as shown in Fig. \ref{fig:multilingual_encoder}. 
Each block had 2 separate Transformer layers. The shared Transformer module contained 2 layers. Therefore, the total number of Transformer layers was also 24.
In this way, we kept all the models in this study to have similar encoder sizes.
For LAMASSU-SPE models, we used three prediction and joint networks, each of which contained two long short-term memory recurrent neural network layers for the prediction network and three linear layers as shown in Equation (\ref{eq:joint_network}) for the joint network.
The vocabulary sizes without $<blank>$ and end of sentence ($<EOS>$) tokens were 4492, 10006, and 4152, for EN, ZH, and DE, respectively. 
For LAMASSU-UNI, after merging repetitive tokens, the vocabulary list contained 14643 tokens excluding $<blank>$ and $<EOS>$.

\begin{table*}[t]
  \caption{Word error rate (WER) and BLEU scores for different source LID schedules. The percentage values denote the percentages of total training steps when we set $\mathbf{g}^j_t$'s to all $1$'s. EN-EN and DE-DE are ASR tasks evaluated using WER, whereas EN-DE, EN-ZH, and DE-EN are ST tasks evaluated with BLEU scores. The best average numbers are in \textbf{bold}.}
  \label{tab:results_lid}
  \setlength{\tabcolsep}{1.8pt}
  \centering
  \resizebox{1.2\columnwidth}{!}{
  \begin{tabular}{l l | c c c | c c c c}
    \toprule
    \multicolumn{2}{c|}{\multirow{2}{*}{methods}} & \multicolumn{3}{c|}{WER for ASR} & \multicolumn{4}{c}{BLEU for ST} \\
    & & EN-EN $\downarrow$ & DE-DE $\downarrow$ & Avg. $\downarrow$ & EN-DE $\uparrow$ & EN-ZH $\uparrow$ & DE-EN $\uparrow$ & Avg. $\uparrow$ \\
    \midrule
    \multirow{4}{*}{LAMASSU-SPE} & 90\% & 15.5 & 29.5 & 22.5 & 18.5 & 19.5 & 17.4 & 18.5 \\
    & 70\% & 14.9 & 28.1 & 21.5 & 19.6 & {20.8} & 18.5 & \textbf{19.6}  \\
    & 50\% & {14.8} & {27.8} & \textbf{21.3} & {19.7} & 20.1 & {18.7} & 19.5  \\
    & 30\% & 15.4 & 28.9 & 22.2 & 18.9 & 18.6 & 17.6 & 18.4  \\
    \midrule
    \multirow{4}{*}{LAMASSU-UNI} & 90\% & 16.5 & 31.0 & 23.8 & 19.3 & 19.1 & 16.7 & 18.4  \\
    & 70\% & 15.7 & 29.6 & 22.7 & {20.0} & 20.6 & 17.7 & 19.4  \\
    & 50\% & {15.1} & {29.0} & \textbf{22.1} & {20.0} & {21.7} & 18.1 & \textbf{19.9}  \\
    & 30\% & 15.3 & {29.0} & 22.2 & 19.7 & 21.0 & {18.2} & 19.6  \\
    \bottomrule
  \end{tabular}
  }
\end{table*}

\begin{table*}[t]
  \caption{WER and BLEU scores for different methods. \emph{\#params} denotes the number of parameters in the model. For LAMASSU-SPE, we include the parameters of all prediction and joint networks. In practice, only one prediction and joint network is used for each target language, and the \emph{\#params} is similar to a single monolingual ASR or bilingual ST model. We calculate the average BLEU score differences between different methods and the baseline monolingual ASR and bilingual ST method as $\Delta Avg. = \mathrm{BLEU}_{method} - \mathrm{BLEU}_{monolingual\ ASR\ \& \ bilingual\ ST}$. Note that target LID for encoder and CTC regularization are applied separately.} 
  \label{tab:results_lamassu}
  \setlength{\tabcolsep}{1.8pt}
  \centering
  \resizebox{1.9\columnwidth}{!}{
  \begin{tabular}{l l | c | c c c | c c c c c}
    \toprule
    \multicolumn{2}{c|}{\multirow{2}{*}{methods}} & \multirow{2}{*}{\#params} & \multicolumn{3}{c|}{WER for ASR} & \multicolumn{5}{c}{BLEU for ST} \\
    & & & EN-EN $\downarrow$ & DE-DE $\downarrow$ & Avg. $\downarrow$ & EN-DE $\uparrow$ & EN-ZH $\uparrow$ & DE-EN $\uparrow$ & Avg. $\uparrow$ & $\Delta$ Avg. $\uparrow$ \\
    \midrule
    \multicolumn{2}{c|}{monolingual ASR \& bilingual ST} & 107M $\times 5$ & 10.3 & 28.1 & 19.2 & 23.1 & 20.1 & 25.0 & 22.7 & 0.0  \\
    \midrule
    \multirow{4}{*}{LAMASSU-SPE} & shared encoder & 146M & 15.6 & 29.2 & 22.4 & 18.4 & 19.4 & 16.8 & 18.2 & -4.5 \\
    & multilingual encoder & 146M & 14.8 & 27.8 & 21.3 & 19.7 & 20.1 & 18.7 & 19.5 & -3.2 \\
    & \hspace{3mm} target LID for encoder &  146M & 14.7 & 26.4 & 20.6 & 20.1 & 21.0 & 18.4 & 19.8 & -2.9 \\
    & \hspace{3mm} CTC regularization & 146M & 14.9 & 27.3 & 21.1 & 19.7 & 21.1 & 18.4 & 19.7 & -3.0  \\
    \midrule
    \multirow{4}{*}{LAMASSU-UNI} & shared encoder & 110M & 17.1 & 31.7 & 24.4 & 18.1 & 19.0 & 15.7 & 17.6 & -5.1 \\
    & multilingual encoder & 110M & {15.1} & {29.0} & 22.1 & {20.0} & {21.7} & 18.1 & 19.9 & -2.8 \\
    & \hspace{3mm} target LID for encoder &  110M & 15.0 & 27.3 & 21.2 & 20.4 & 21.7 & 18.5 & 20.2 & -2.5 \\
    & \hspace{3mm} CTC regularization & 110M & 12.6 & 25.3 & \textbf{19.0} & 23.6 & 25.5 & 23.3 & {24.1} & \textbf{+1.4} \\
    \bottomrule
  \end{tabular}
  }
\end{table*}

To train the models, we set the numbers of total training steps proportional to the sizes of training sets. For monolingual ASR and bilingual ST models, we trained them for 100K steps. For LAMASSU models, the training steps were 600K since the sizes of their training labels were six times those for monolingual ASR and bilingual ST. This way, each training sample was used for the same number of times in different training setups. We used 16 graphics processing units (GPUs) and set the batch size to 400K for all experiments.

\section{Evaluation Results}
\label{sec:eval}

\subsection{Source LID Scheduling for Clustered Multilingual Encoder}
\label{ssec:eval_lid}
Table \ref{tab:results_lid} shows the comparisons among different source LID scheduling configurations. 
50\% and 70\% performed close and 50\% was better in general. Therefore, in most experiments, we used 50\% as the percentage of total training steps when we set $\mathbf{g}^j_t$'s to all $1$'s. However, for experiments with CTC regularization, we used 70\% since it obtained slightly better results than 50\%.
The DE WERs were relatively high. This is because our DE training data was mismatched with the common voice DE test set and we only used a subset of the training set. Nevertheless, this data mismatch does not influence the comparison between baseline models and LAMASSU.

\subsection{LAMASSU}
\label{ssec:eval_lamassu}
Table \ref{tab:results_lamassu} shows the comparisons among different methods. 
The clustered multilingual encoder improved the performances of both LAMASSU-SPE and LAMASSU-UNI over shared encoders by about 2 BLEU scores on average.
Feeding target LID to the encoder of LAMASSU yielded consistent improvements for both LAMASSU-SPE and LAMASSU-UNI. 
With CTC regularization, LAMASSU-UNI outperformed the baseline monolingual ASR and bilingual ST models. 
Note that CTC regularization is especially useful for small datasets like the one in this paper. It can also be used for baseline bilingual ST models. We leave the investigation of the influence of CTC and the comparison between different methods on production-scale datasets as future work.
We emphasize that although LAMASSU-SPE did not gain a large improvement with the proposed methods, it still has advantages over LAMASSU-UNI in the flexibility of deployment and adding new target languages. 
All the models in this study are streaming. The average proportion, average lagging, and differentiable average lagging values are close to 0.61, 841ms, and 834ms, respectively. We show the streaming capability of LAMASSU with the online demo \footref{cvpr_video}.

\section{Conclusion}
\label{sec:conclusion}
We proposed LAMASSU, the first streaming language-agnostic multilingual ASR and many-to-many ST model using neural transducers. Specifically for the transducer model architecture, we proposed four methods, LAMASSU-UNI, a clustered multilingual encoder, target LID for encoder, and CTC regularization.
In addition to the novel clustered multilingual encoder and target LID for encoder methods, LAMASSU-UNI is the first multilingual ST model using a single prediction and joint network in neural transducers, and auxiliary CTC loss is used for transducer-based ST tasks for the first time.
With the proposed methods, our LAMASSU-UNI model reached the performances of monolingual ASR and bilingual ST models, while dramatically reducing the model size.

\bibliographystyle{IEEEtran}
\bibliography{mybib}

\end{document}